 \documentclass[pmlr,twocolumn,10pt]{include/ml4h/jmlr} %

\usepackage{booktabs}
\usepackage{siunitx}

\usepackage{microtype}
\usepackage{graphicx}
\usepackage{booktabs} %

\usepackage{titletoc}
\usepackage{amsmath}
\usepackage{amssymb}
\usepackage{hyperref}

\usepackage{multirow}
\usepackage{array}
\usepackage{multicol}
\usepackage{makecell}
\usepackage{lipsum}
\usepackage{float}

\usepackage[utf8]{inputenc}
\usepackage{microtype}
\usepackage{graphicx}
\usepackage{booktabs} %
\usepackage{titletoc}
\usepackage{hyperref}

\usepackage{amsmath}
\usepackage{amssymb}
\usepackage{bbm} 
\usepackage{bm}
\usepackage{verbatim}
\usepackage{float}
\usepackage{color,soul}
\usepackage{enumitem}
\usepackage{mathtools}
\usepackage{hhline}
\usepackage[title]{appendix}
\usepackage[nameinlink]{cleveref} %
\usepackage{tikz}
\usepackage{wrapfig,booktabs}
\usepackage{xspace}
\usepackage{cancel}
\usepackage{sidecap}

\graphicspath{ {../figures/} }

\DeclarePairedDelimiterX{\infdivx}[2]{(}{)}{%
  #1\;\delimsize\|\;#2%
}

\newcommand{\kld}{KL divergence\xspace}

\newcommand{\real}{\mathbb{R}}

\crefname{appsec}{appendix}{appendices}
\Crefname{appsec}{Appendix}{Appendices}

\definecolor{mydarkblue}{rgb}{0,0.08,0.45}
\hypersetup{ %
    pdftitle={},
    pdfauthor={},
    pdfsubject={Proceedings of the International Conference on Machine Learning 2020},
    pdfkeywords={},
    pdfborder=0 0 0,
    pdfpagemode=UseNone,
    colorlinks=true,
    linkcolor=mydarkblue,
    citecolor=mydarkblue,
    filecolor=mydarkblue,
    urlcolor=mydarkblue,
    pdfview=FitH
}

\usetikzlibrary{shapes.geometric, arrows, bayesnet, calc, positioning}

\newcommand{\calF}{\mathcal{F}}

\newcommand{\calN}{\mathcal{N}}

\newcommand{\closer}[3]{{\kern-#1ex{#2}\kern-#3ex}}

\newcommand{\DKL}{D_{\text{KL}}\infdivx}

\DeclareMathOperator*{\argmax}{arg\,max}

\mathchardef\mhyphen="2D

\usepackage{titletoc}
\usepackage{algorithm}
\usepackage{algorithmic}
\usepackage{colortbl}

\definecolor{azure}{rgb}{0.0, 0.5, 1.0}
\definecolor{airforceblue}{rgb}{0.36, 0.54, 0.66}
\definecolor{darkgreen}{rgb}{0.0, 0.2, 0.13}

\newcommand\defines{\,\dot{=}\,}

\newcommand{\vbar}{\,|\,}

\newcommand{\calY}{\mathcal{Y}}
\newcommand{\calD}{\mathcal{D}}

\newcommand{\calQ}{\mathcal{Q}}

\usepackage{standalone}
\usepackage{tikz}
\usepackage{pgfplots}
\usepackage{pgf}
\usetikzlibrary{calc}
\usetikzlibrary{positioning}
\usetikzlibrary{angles,quotes}
\usetikzlibrary{backgrounds}
\usetikzlibrary{fit}
\usetikzlibrary{arrows}
\usetikzlibrary{arrows.meta}
\usetikzlibrary{shapes.symbols}
\usetikzlibrary{shadings}
\usetikzlibrary{shapes}
\usetikzlibrary{fadings}
\usetikzlibrary{bayesnet}
\usetikzlibrary{matrix}
\usetikzlibrary{plotmarks}
\usetikzlibrary{intersections}
\usetikzlibrary{pgfplots.fillbetween}
\pgfplotsset{compat=1.14}
\usepackage{siunitx}

\usepackage{colortbl}
\definecolor{mediumgray}{gray}{0.7}
\definecolor{lightgray}{gray}{0.85}
\definecolor{lightlightgray}{gray}{0.9}
\definecolor{C1}{HTML}{1F77B4}
\definecolor{C2}{HTML}{FF7F0E}
\definecolor{C3}{HTML}{2CA02C}
\definecolor{C4}{HTML}{D62728}
\definecolor{C5}{HTML}{9467BD}
\colorlet{C1light}{C1!70!white}
\colorlet{C2light}{C2!70!white}
\colorlet{C3light}{C3!70!white}
\colorlet{C4light}{C4!70!white}
\colorlet{C5light}{C5!70!white}
\colorlet{C1lighter}{C1!50!white}
\colorlet{C2lighter}{C2!50!white}
\colorlet{C3lighter}{C3!50!white}
\colorlet{C4lighter}{C4!50!white}
\colorlet{C5lighter}{C5!50!white}
\colorlet{C1vlight}{C1!20!white}
\colorlet{C2vlight}{C2!20!white}
\colorlet{C3vlight}{C3!20!white}
\colorlet{C4vlight}{C4!20!white}
\colorlet{C5vlight}{C5!20!white}
\colorlet{linkcolor}{violet}
\usepackage[algo2e]{algorithm2e}

\newcommand{\track}[1]{{\color{black} #1}}

\jmlrvolume{LEAVE UNSET}
\jmlryear{2023}
\jmlrsubmitted{LEAVE UNSET}
\jmlrpublished{LEAVE UNSET}
\jmlrworkshop{Machine Learning for Health (ML4H) 2023} %

\title[Informative Priors Improve the Reliability of Multimodal Clinical Data Classification]{Informative Priors Improve the Reliability of\\Multimodal Clinical Data Classification}
\author{%
\Name{L. Julian Lechuga Lopez}$^{1,2}$ \Email{leopoldo.lechuga@nyu.edu} \\
\Name{Tim G. J. Rudner}$^{2}$ \Email{tim.rudner@nyu.edu} \\
\Name{Farah E. Shamout}$^{1,2}$ \Email{farah.shamout@nyu.edu} \\
\addr $^1$NYU Abu Dhabi, AD, UAE \\
\addr $^2$New York University, NY, USA\\
}

\begin{document}

\maketitle

\begin{abstract}
Machine learning-aided clinical decision support has the potential to significantly improve patient care.  However, existing efforts in this domain for principled quantification of uncertainty have largely been limited to applications of ad-hoc solutions that do not consistently improve reliability. In this work, we consider stochastic neural networks and design a tailor-made multimodal data-driven (\textsc{m2d2}) prior distribution over network parameters. We use simple and scalable Gaussian mean-field variational inference to train a Bayesian neural network using the \textsc{m2d2} prior. We train and evaluate the proposed approach using clinical time-series data in MIMIC-IV and corresponding chest X-ray images in MIMIC-CXR for the classification of acute care conditions. Our empirical results show that the proposed method produces a more reliable predictive model compared to deterministic and Bayesian neural network baselines.
\end{abstract}


\begin{figure*}[t!]
    \centering
    \vspace*{-15pt}\hspace*{0.7cm}
    \includegraphics[width=0.95\textwidth]{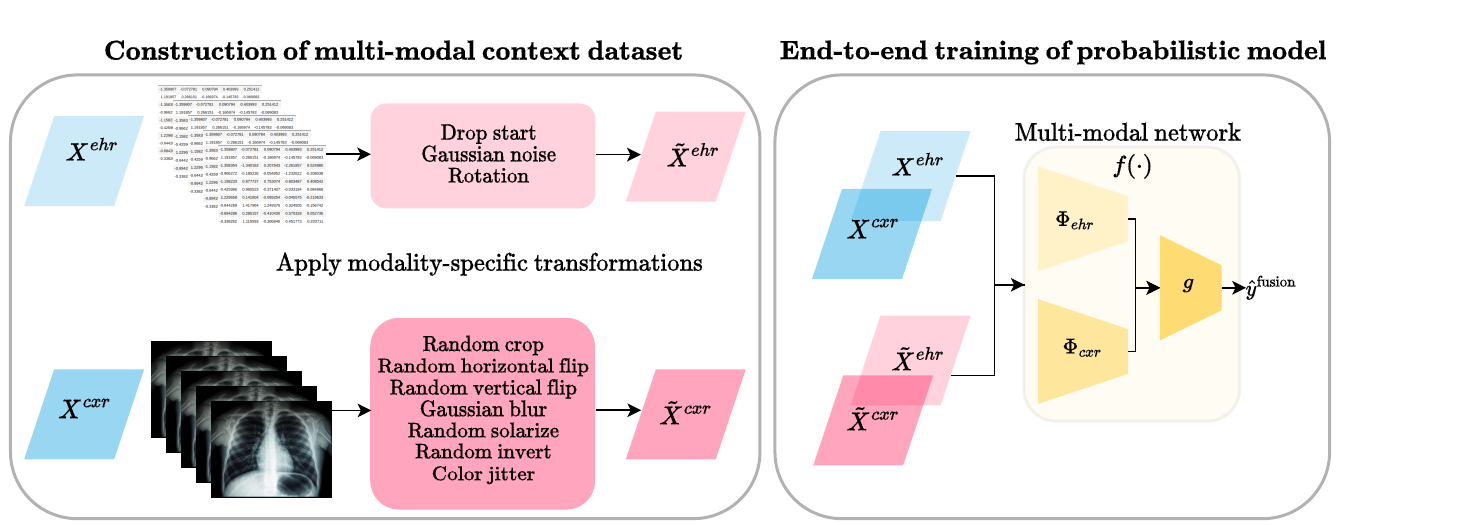}
    \vspace*{-10pt}
    \caption{
        \small\textbf{Overview of model training.}
        \textit{Left:} We construct a multimodal context dataset by applying modality-specific transformations to clinical time series data and chest X-ray images, resulting in a tailored distribution shift.
        \textit{Right:} We train the multimodal neural network end-to-end with the training set and the constructed context dataset.}
    \label{fig:fusion_encoder}
    \vspace*{-10pt}
\end{figure*}

\section{Introduction}
\label{sec:intro}

Trustworthy machine learning in healthcare requires robust uncertainty quantification \citep{begoli2019, gruber2023sources}, considering the safety-critical nature of clinical practice. Sources of uncertainty can be due to model parameters, noise and bias of the calibration data, or deployment of the model in an out-of-distribution scenario \citep{miller2014advanced}.

Unfortunately, the literature in machine learning for healthcare has largely neglected developing tailored solutions for improved uncertainty quantification \citep{kompa2021second}, perhaps due to the limited underlying theory on how to best adapt predictive uncertainty in clinical tasks \citep{begoli2019}. 
Other challenges include the complexity of scaling uncertainty quantification in real-time clinical systems, limited empirical evaluation of different methods due to the lack of well-constructed priors by medical experts \citep{zou2023review}, and the high prevalence of data shifts in real-world clinical applications that can negatively affect predictive performance \citep{ovadia2019can, xia2022benchmarking}, further emphasizing the need for better uncertainty in predictive models.

Additionally, despite the recent proliferation of multimodal learning, existing work on uncertainty quantification in healthcare has mainly been studied in the unimodal setting, with a particular focus on medical imaging applications \citep{gawlikowski2021survey}.
This includes brain tumor segmentation \citep{jungo2018towards}, skin lesion segmentation \citep{devries2018leveraging}, and diabetic retinopathy detection tasks \citep{filos2019systematic,Band2021benchmarking,nado2022uncertainty}, among others.
Hence, effective quantification of predictive uncertainty in the context of multimodal clinical problems remains a challenging and unsolved task~\citep{Plex22}.

We propose a multimodal data-driven (\textsc{m2d2}) prior distribution over neural network parameters to improve uncertainty quantification in multimodal fusion of chest X-ray images and clinical time series data.
We evaluate the use of effective priors on the two unimodal components of the multimodal fusion network: an image-based convolutional neural network and a recurrent neural network for clinical time series.
In summary, we make the following contributions:\vspace*{-6pt}
\begin{enumerate}[leftmargin=15pt]
\setlength\itemsep{-2pt}
    \item We design a multimodal data-driven (\textsc{m2d2}) prior distribution over neural network parameters that places high probability density on desired predictive functions.
    \item We evaluate the method on large publicly-available multimodal datasets: MIMIC-IV and MIMIC-CXR \citep{mimiccxr, mimiciv}, for the classification of acute care conditions assigned to patient stays in the intensive care unit.
    \item Our findings illustrate an increase in predictive performance and improved reliability in uncertainty-aware selective prediction.
\end{enumerate}

\vspace{-1mm}
\section{Related Work}
\label{sec:background}

\subsection{Multimodal Learning in Healthcare}

Multimodal learning in healthcare seeks to exploit complementary information from different data modalities to enhance the predictive capabilities of learning models. There are different approaches for leveraging information across different data modalities, with the most popular paradigm being multimodal fusion \citep{huang2020fusion}.
For example, \cite{zhang2020advances} and \cite{calhoun2016multimodal} investigated different methods for fusion segmentation and quantification in neuroimaging by leveraging different imaging modalities in the same data pipeline.
Another recent study focused on the development of smart healthcare applications by merging multimodal signals collected from different types of medical sensors \citep{muhammad2021comprehensive}. 
Other studies also show improved predictive performance when using multiple modalities in prognostic tasks in patients with COVID-19 \citep{shamout2021artificial, jiao2021prognostication}. 

Despite the promise of multimodal learning in healthcare, research in reliable uncertainty quantification applications in the multimodal setting is currently limited.
There is no generalized use of uncertainty quantification methods that address increased data distribution shifts and deal with multiple modalities simultaneously \citep{foundationsmultimodal}. 

\vspace*{-8pt}
\subsection{Variational Inference in Neural Networks}

We consider a stochastic neural network $f(\cdot \,; \Theta)$, defined in terms of stochastic parameters \mbox{$\Theta \in \real^{P}$}.
For an observation model \mbox{$p_{Y | X, \Theta}$} and a prior distribution over parameters $p_{\Theta}$, Bayesian inference provides a mathematical formalism for finding the posterior distribution over parameters given the observed data, $p_{\Theta | \calD}$~\citep{mckay1992practical,neal1996bayesian}.
However, since neural networks are non-linear in their parameters, exact inference over the stochastic network parameters is analytically intractable.

\begin{figure*}[h]
    \centering
       \vspace*{-5pt}
       \includegraphics[width=0.8\textwidth]{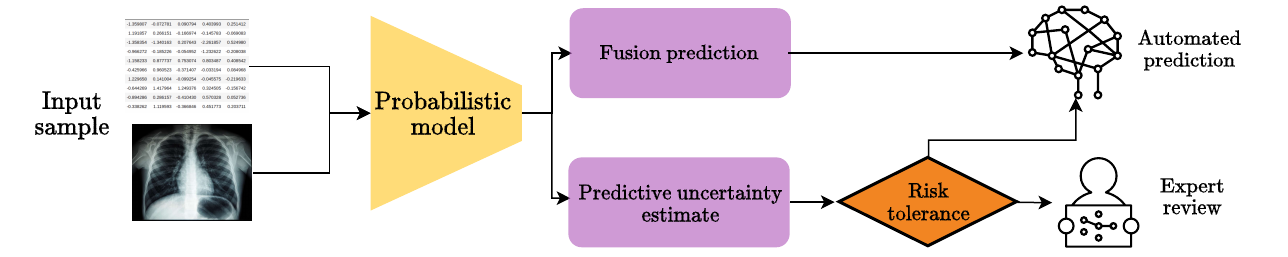}
    \caption{\small
        \textbf{Selective prediction algorithm.} For each input sample, we obtain a prediction and an estimate of the model's uncertainty for that specific data point. If the uncertainty estimate is higher than the selected risk tolerance, then the sample will be sent to an expert for further review and classification. Otherwise, the sample is processed by the learning model for automated prediction.
    }
    \label{fig:selective_prediction}
    \vspace*{-12pt}
\end{figure*}

Variational inference is an approach that seeks to avoid this intractability by framing posterior inference as finding an approximation $q_{\Theta}$ to the posterior $p_{\Theta | \calD}$ via the variational optimization problem:\vspace*{-5pt}
\begin{align*}
    \min\nolimits_{q_{\Theta} \in \calQ_{\Theta}} \DKL{q_{\Theta}}{p_{\Theta | \calD}}
    \Longleftrightarrow
    \max\nolimits_{q_{\Theta} \in \calQ_{\Theta}}  \calF(q_{\Theta}) ,
\end{align*}\\[-15pt]
where $\calF(q_{\Theta})$ is the variational objective\vspace*{-5pt}
\begin{align}
    \calF(q_{\Theta})
    \hspace*{-1pt}\defines\hspace*{-1pt}
    \mathbb{E}_{q_{\Theta}}[\log p(y_{\calD} \vbar x_{\calD}, \Theta) ] \hspace*{-1pt}-\hspace*{-1pt} \DKL{q_{\Theta}}{p_{\Theta}} ,
    \label{eq:elbo}
\end{align}\\[-15pt]
$\calQ_{\Theta}$ is a variational family of distributions~\citep{wainwright2008vi}, and $(x_{\calD}, y_{\calD})$ are the training data.
One particularly simple type of variational inference is Gaussian mean-field variational inference \citep{blundell2015mfvi,graves2011practical}, where the posterior distribution over network parameters is approximated by a Gaussian distribution with a diagonal covariance matrix.
This method enables stochastic optimization and can be scaled to large neural networks~\citep{hoffman2013svi}.
However, Gaussian mean-field variational inference has been shown to underperform with deterministic neural networks when uninformative, standard Gaussian priors are used~\citep{ovadia2019uncertainty,fsvi}.

To improve the performance, we extend the approach presented in \citet{rudner2023fseb} to stochastic neural networks, construct a data-driven prior distribution from multimodal input data, and use this prior for Gaussian mean-field variational inference to improve the performance of neural networks for multimodal clinical prediction tasks.

\section{Constructing Data-Driven Priors for Models with Multimodal Input Data}
\label{sec:method}
We consider a supervised multimodal fusion task on data \mbox{$\mathcal{D}\doteq\{(x^{1}_n, x^{2}_n,{y}_n^{\textrm{fusion}})\}^{\mathit{N}}_{n=1}=(X^{1}_\mathcal{D},X^{2}_\mathcal{D},{Y}_\mathcal{D})$}.
As shown in Figure~\ref{fig:fusion_encoder}, we consider the first modality to be clinical time series data extracted from electronic health records, denoted by $X^{\textrm{ehr}}$, and the second to be chest X-ray images, denoted by $X^{\textrm{cxr}}$.
For a given sample $(x^{\textrm{ehr}}, x^{\textrm{cxr}})$, the two modalities are processed by encoders $\Phi_{\textrm{ehr}}$ and $\Phi_{\textrm{cxr}}$ respectively, and their concatenated feature representations are further processed by a classifier $g(\cdot)$ and activation function to compute the fusion prediction $\hat{y}^{\textrm{fusion}}$.
The loss is computed based on the predictions and ground truth labels $y^{\textrm{fusion}} \in \calY$, where $\mathcal{Y}\subseteq\{0,1\}^\mathit{Q}$, with $\mathit{Q}>1$ for multi-label classification. 

\vspace*{-3pt}
\subsection{\hspace*{-3.5pt}Informative Priors for Multimodal Data}
\label{sec:context}
\vspace*{-3pt}

One of the key components in defining the probabilistic model for our uncertainty quantification method is the definition of a sensible and explainable prior distribution. 
In this work, we construct a prior distribution over parameters that places high probability density on parameter values that induce predictive functions that have high uncertainty on input points that are meaningfully different from the training data.
To do this, we build on the approach proposed in \citet{rudner2023fseb} and use information about the two input modalities to construct a data-driven prior that can help find an approximate posterior distribution with desirable properties (e.g., an induced predictive distribution with reliable uncertainty estimation).
More specifically, we construct a data-driven prior over some set of model parameters $\Psi$ and condition it on a set of context points $\tilde{X}$, that is $p(\psi | \tilde{x})$.
In \Cref{appsec:variational_objective}, we show that we can derive a tractable variational objective using this prior.
The objective is given by \Cref{eq:final_objective}.

To construct a meaningful prior, we need to specify a distribution over the set of context points, $p_{\tilde{X}}$.
We design a multimodal prior by letting $\tilde{X}$ be a set of randomly generated multimodal input points $(\tilde{X}^{\textrm{ehr}}, \tilde{X}^{\textrm{cxr}})$ designed to be distinct from the training data.
For the clinical time series data, we construct $\tilde{X}^{\textrm{ehr}}$ by applying three transformations to the original time series: drop start, Gaussian noise, and inversion (i.e, for each $x_i$ in $1,...,n$, $x_1=x_n$, $x_2=x_{n-1}$, $x_3=x_{n-2}$, etc).
For the chest X-ray images, we construct $\tilde{X}^{\textrm{cxr}}$ by applying seven transformations representative of perturbations that exist in real-world medical settings to the imaging data: random crop, random horizontal and vertical flip, Gaussian blur, random solarize, random invert and color jitter.

Hence, this context set encompasses  distributionally shifted points,  where we want the model's uncertainty to be higher.
\begin{table*}[ht!]
\vspace*{-15pt}
\caption{
    \small \textbf{Performance results.}
    We summarize the results on the test set for the baselines and our stochastic model, including 95\% confidence intervals computed via bootstrapping.
    Higher values are better for all metrics.
    }
\label{table:main_results}
\vspace*{-8pt}
    \centering
    \resizebox{1.0\linewidth}{!}{
    \begin{tabular}{lcccccc}
    \toprule
    \textbf{Model (MedFuse)} & \textbf{AUROC} & \textbf{AUPRC} & \textbf{\makecell{Selective AUROC}} & \textbf{\makecell{Selective AUPRC}}
    \\
    \midrule

    \textbf{Deterministic}~\citep{medfuse} & 0.726 (0.718, 0.733) & 0.503 (0.493, 0.517) & 0.724 (0.715, 0.735) & 0.439 (0.429, 0.455)
    \\
    \textbf{Bayesian} (standard prior) & 0.729 (0.722, 0.736) & 0.507 (0.497, 0.521) & \cellcolor[gray]{0.9}\textbf{0.748} (0.739, 0.758) & 0.448 (0.437, 0.467)
    \\
     \textbf{Bayesian} (\textsc{m2d2} prior) (\textbf{Ours}) & \cellcolor[gray]{0.9}\textbf{0.735} (0.728, 0.742) & \cellcolor[gray]{0.9}\textbf{0.514} (0.504, 0.528) & \cellcolor[gray]{0.9}\textbf{0.748} (0.738, 0.760) & \cellcolor[gray]{0.9}\textbf{0.452} (0.441, 0.472) \\ %
    \bottomrule
    \end{tabular}
    }
    \vspace*{-5pt}
\end{table*}

\vspace{-1mm}
\section{Empirical Evaluation}
\label{sec:experiments}

To evaluate the proposed approach, we combine clinical time series data from MIMIC-IV \citep{mimiciv} and chest X-ray images from MIMIC-CXR \citep{mimiccxr} collected during the same patient stay in the intensive care unit for multi-label classification of acute care conditions.

\vspace*{-5pt}
\subsection{Experimental Setup}

We follow the pre-processing steps and use the same neural network architecture (MedFuse) as \citet{medfuse}.
$\Phi_{\textrm{ehr}}$ is a two-layer LSTM network \citep{lstm}, $\Phi_{\textrm{cxr}}$ is a ResNet-34 \citep{resnet50}, $g(\cdot)$ is a fully connected layer, and $\hat{y}^{\textrm{fusion}}$ are the class probabilities obtained by applying a sigmoid function to $g$.
We use the paired dataset, such that each sample contains both modalities (i.e., there are no missing modalities).
Hence, the training, validation, and test sets consisted of $7756$, $877$, and $2161$ samples, respectively.
We construct the context dataset using the training set.

We train the multimodal network for 400 epochs using the loss presented in \Cref{equation:loss}, with the Adam optimizer, a batch size of $16$, and a learning rate of $2\times10^{-4}$.
Further details on the experimental setup and grid-based hyperparameter tuning can be found in \Cref{app:experimental-setup}.

\vspace*{-5pt}
\subsection{Evaluation Metrics}

We evaluate the overall performance of the models on the test set using the AUROC and Area Under the Precision-Recall curve (AUPRC) \citep{medfuse}.

In addition, we compute selective prediction evaluation metrics to better assess models' predictive uncertainty. 
As shown in \Cref{fig:selective_prediction}, selective prediction modifies the standard prediction pipeline by introducing a ``reject option'', $\bot$, via a gating mechanism defined by selection function $s:\mathcal{X}\to\mathbb{R}$ that determines whether a prediction should be made for a given input point $x\in\mathcal{X}$  \citep{el2010foundations}.
For rejection threshold $\tau$, with $s$ representing the entropy of $x$, the prediction model is given by\vspace*{-7pt}
\begin{align}
    (p(y\,|\,\cdot,\mathbf{\theta};f),s)(x) =  
    \begin{cases}
          p(y\,|\,x,\mathbf{\theta};f), & \text{if}\ s\le \tau \\
          \bot, & \text{otherwise}
    \end{cases}
\end{align}\\[-18pt]

To evaluate the predictive performance of a prediction model $(p(y\,|\,\cdot,\mathbf{\theta};f),s)(x)$ with a single label, we compute the AUROC and AUPRC over rejection thresholds $\tau=0\%,...,99\%$.
We then average the metrics across all thresholds, yielding selective prediction AUROC and AUPRC scores that explicitly incorporate both a model's predictive performance and its predictive uncertainty.
For our multi-label classification task, we report the average selective prediction scores across all 25 labels.

\vspace*{-5pt}
\subsection{Results}
\label{sec:results}

\Cref{table:main_results} summarizes the performance results on the test set.
Additional per-label results are shown in Appendix~\ref{appsec:results}.
The Bayesian neural network with an \textsc{m2d2} prior achieves a better AUROC and AUPRC of 0.735 and 0.514, respectively, compared with 0.726 and 0.503 by the deterministic model.
It also achieves a higher selective AUROC and AUPRC of 0.748 and 0.452, respectively, compared with 0.724 and 0.439 by the deterministic model.
Our proposed approach achieves a comparable selective AUROC to when using a standard prior.

\track{
We also observe a decrease in selective AUPRC compared to the 0\%-rejection AUPRC.
This can occur when a model is poorly calibrated: When the AUPRC for any rejection threshold is below the 0\%-rejection score, the selective AUPRC can be lower than the 0\%-rejection score. 
}
Overall, the selective prediction scores reflect the model's ability to identify samples that are more likely to be misclassified and should be reviewed by a clinician, and as such, are valuable in assessing model reliability in clinical settings.

\vspace*{-5pt}
\section{Conclusion}
\label{sec:conclusion}

We designed a multimodal data-driven (\textsc{m2d2}) prior to improve the reliability of multimodal fusion of clinical time series data and chest X-ray images.
We demonstrated that Bayesian neural networks with such a prior achieve better performance, in terms of AUROC, AUPRC, and selective prediction scores, than deterministic models. 
For future work, we aim to evaluate the proposed approach in settings of missing modalities, on additional tasks, such as in-hospital mortality prediction, and other multi-modal datasets. 

\section{Acknowledgements}
\label{sec:acknowledgements}
This research was carried out on the High Performance Computing resources at New York University Abu Dhabi. We would also like to thank postdoctoral associate Dr. Alejandro Guerra Manzanares for the helpful discussions and support in refactoring code from Pytorch to JAX.

\bibliography{references}

\begin{thebibliography}{42}
\providecommand{\natexlab}[1]{#1}
\providecommand{\url}[1]{\texttt{#1}}
\expandafter\ifx\csname urlstyle\endcsname\relax
  \providecommand{\doi}[1]{doi: #1}\else
  \providecommand{\doi}{doi: \begingroup \urlstyle{rm}\Url}\fi

\bibitem[Band et~al.(2021)Band, Rudner, Feng, Filos, Nado, Dusenberry, Jerfel,
  Tran, and Gal]{Band2021benchmarking}
Neil Band, Tim G.~J. Rudner, Qixuan Feng, Angelos Filos, Zachary Nado,
  Michael~W. Dusenberry, Ghassen Jerfel, Dustin Tran, and Yarin Gal.
\newblock Benchmarking {B}ayesian deep learning on diabetic retinopathy
  detection tasks.
\newblock 2021.

\bibitem[Begoli et~al.(2019)Begoli, Bhattacharya, and Kusnezov]{begoli2019}
Edmon Begoli, Tanmoy Bhattacharya, and Dimitri Kusnezov.
\newblock The need for uncertainty quantification in machine-assisted medical
  decision making.
\newblock \emph{Nature Machine Intelligence}, 1\penalty0 (1):\penalty0 20--23,
  2019.

\bibitem[Blundell et~al.(2015)Blundell, Cornebise, Kavukcuoglu, and
  Wierstra]{blundell2015mfvi}
Charles Blundell, Julien Cornebise, Koray Kavukcuoglu, and Daan Wierstra.
\newblock Weight uncertainty in neural networks.
\newblock volume~37 of \emph{Proceedings of Machine Learning Research}, pages
  1613--1622, Lille, France, 07--09 Jul 2015. PMLR.

\bibitem[Bradbury et~al.(2018)Bradbury, Frostig, Hawkins, Johnson, Leary,
  Maclaurin, Necula, Paszke, Vander{P}las, Wanderman-{M}ilne, and Zhang]{jax}
James Bradbury, Roy Frostig, Peter Hawkins, Matthew~James Johnson, Chris Leary,
  Dougal Maclaurin, George Necula, Adam Paszke, Jake Vander{P}las, Skye
  Wanderman-{M}ilne, and Qiao Zhang.
\newblock {JAX}: composable transformations of {P}ython+{N}um{P}y programs,
  2018.

\bibitem[Calhoun and Sui(2016)]{calhoun2016multimodal}
Vince~D Calhoun and Jing Sui.
\newblock Multimodal fusion of brain imaging data: a key to finding the missing
  link (s) in complex mental illness.
\newblock \emph{Biological psychiatry: cognitive neuroscience and
  neuroimaging}, 1\penalty0 (3):\penalty0 230--244, 2016.

\bibitem[DeVries and Taylor(2018)]{devries2018leveraging}
Terrance DeVries and Graham~W Taylor.
\newblock Leveraging uncertainty estimates for predicting segmentation quality.
\newblock \emph{arXiv preprint arXiv:1807.00502}, 2018.

\bibitem[El-Yaniv et~al.(2010)]{el2010foundations}
Ran El-Yaniv et~al.
\newblock On the foundations of noise-free selective classification.
\newblock \emph{Journal of Machine Learning Research}, 11\penalty0 (5), 2010.

\bibitem[Filos et~al.(2019)Filos, Farquhar, Gomez, Rudner, Kenton, Smith,
  Alizadeh, De~Kroon, and Gal]{filos2019systematic}
Angelos Filos, Sebastian Farquhar, Aidan~N Gomez, Tim~GJ Rudner, Zachary
  Kenton, Lewis Smith, Milad Alizadeh, Arnoud De~Kroon, and Yarin Gal.
\newblock A systematic comparison of {B}ayesian deep learning robustness in
  diabetic retinopathy tasks.
\newblock \emph{arXiv preprint arXiv:1912.10481}, 2019.

\bibitem[Gawlikowski et~al.(2021)Gawlikowski, Tassi, Ali, Lee, Humt, Feng,
  Kruspe, Triebel, Jung, Roscher, et~al.]{gawlikowski2021survey}
Jakob Gawlikowski, Cedrique Rovile~Njieutcheu Tassi, Mohsin Ali, Jongseok Lee,
  Matthias Humt, Jianxiang Feng, Anna Kruspe, Rudolph Triebel, Peter Jung,
  Ribana Roscher, et~al.
\newblock A survey of uncertainty in deep neural networks.
\newblock \emph{arXiv preprint arXiv:2107.03342}, 2021.

\bibitem[Good(1952)]{good1952rational}
Irving~John Good.
\newblock Rational decisions.
\newblock \emph{Journal of the Royal Statistical Society: Series B
  (Methodological)}, 14\penalty0 (1):\penalty0 107--114, 1952.

\bibitem[Graves(2011)]{graves2011practical}
Alex Graves.
\newblock Practical variational inference for neural networks.
\newblock \emph{Advances in neural information processing systems}, 24, 2011.

\bibitem[Gruber et~al.(2023)Gruber, Schenk, Schierholz, Kreuter, and
  Kauermann]{gruber2023sources}
Cornelia Gruber, Patrick~Oliver Schenk, Malte Schierholz, Frauke Kreuter, and
  Göran Kauermann.
\newblock Sources of uncertainty in machine learning -- a statisticians' view,
  2023.

\bibitem[Hayat et~al.(2022)Hayat, Geras, and Shamout]{medfuse}
Nasir Hayat, Krzysztof~J. Geras, and Farah~E. Shamout.
\newblock Medfuse: Multi-modal fusion with clinical time-series data and chest
  x-ray images.
\newblock In \emph{Proceedings of the 7th Machine Learning for Healthcare
  Conference}, volume 182 of \emph{Proceedings of Machine Learning Research},
  pages 479--503. PMLR, 05--06 Aug 2022.

\bibitem[He et~al.(2015)He, Zhang, Ren, and Sun]{resnet50}
Kaiming He, Xiangyu Zhang, Shaoqing Ren, and Jian Sun.
\newblock Deep residual learning for image recognition, 2015.

\bibitem[Hochreiter and Schmidhuber(1997)]{lstm}
Sepp Hochreiter and J{\"u}rgen Schmidhuber.
\newblock Long short-term memory.
\newblock \emph{Neural computation}, 9\penalty0 (8):\penalty0 1735--1780, 1997.

\bibitem[Hoffman et~al.(2013)Hoffman, Blei, Wang, and Paisley]{hoffman2013svi}
Matthew~D. Hoffman, David~M. Blei, Chong Wang, and John Paisley.
\newblock Stochastic variational inference.
\newblock \emph{Journal of Machine Learning Research}, 14\penalty0
  (1):\penalty0 1303--1347, May 2013.
\newblock ISSN 1532-4435.

\bibitem[Huang et~al.(2020)Huang, Pareek, Seyyedi, Banerjee, and
  Lungren]{huang2020fusion}
Shih-Cheng Huang, Anuj Pareek, Saeed Seyyedi, Imon Banerjee, and Matthew~P
  Lungren.
\newblock Fusion of medical imaging and electronic health records using deep
  learning: a systematic review and implementation guidelines.
\newblock \emph{NPJ digital medicine}, 3\penalty0 (1):\penalty0 136, 2020.

\bibitem[Jiao et~al.(2021)Jiao, Choi, Halsey, Tran, Hsieh, Wang, Eweje, Wang,
  Chang, Wu, et~al.]{jiao2021prognostication}
Zhicheng Jiao, Ji~Whae Choi, Kasey Halsey, Thi My~Linh Tran, Ben Hsieh, Dongcui
  Wang, Feyisope Eweje, Robin Wang, Ken Chang, Jing Wu, et~al.
\newblock Prognostication of patients with covid-19 using artificial
  intelligence based on chest x-rays and clinical data: a retrospective study.
\newblock \emph{The Lancet Digital Health}, 3\penalty0 (5):\penalty0
  e286--e294, 2021.

\bibitem[Johnson et~al.(2021)Johnson, Bulgarelli, Pollard, Celi, Mark, and
  Horng~IV]{mimiciv}
Alistair Johnson, Lucas Bulgarelli, Tom Pollard, Leo~Anthony Celi, Roger Mark,
  and S~Horng~IV.
\newblock Mimic-iv-ed.
\newblock \emph{PhysioNet}, 2021.

\bibitem[Johnson et~al.(2019)Johnson, Pollard, Berkowitz, Greenbaum, Lungren,
  Deng, Mark, and Horng]{mimiccxr}
Alistair~EW Johnson, Tom~J Pollard, Seth~J Berkowitz, Nathaniel~R Greenbaum,
  Matthew~P Lungren, Chih-ying Deng, Roger~G Mark, and Steven Horng.
\newblock Mimic-cxr, a de-identified publicly available database of chest
  radiographs with free-text reports.
\newblock \emph{Scientific data}, 6\penalty0 (1):\penalty0 317, 2019.

\bibitem[Jungo et~al.(2018)Jungo, McKinley, Meier, Knecht, Vera,
  P{\'e}rez-Beteta, Molina-Garc{\'\i}a, P{\'e}rez-Garc{\'\i}a, Wiest, and
  Reyes]{jungo2018towards}
Alain Jungo, Richard McKinley, Raphael Meier, Urspeter Knecht, Luis Vera,
  Juli{\'a}n P{\'e}rez-Beteta, David Molina-Garc{\'\i}a, V{\'\i}ctor~M
  P{\'e}rez-Garc{\'\i}a, Roland Wiest, and Mauricio Reyes.
\newblock Towards uncertainty-assisted brain tumor segmentation and survival
  prediction.
\newblock In \emph{Brainlesion: Glioma, Multiple Sclerosis, Stroke and
  Traumatic Brain Injuries: Third International Workshop, BrainLes 2017, Held
  in Conjunction with MICCAI 2017, Quebec City, QC, Canada, September 14, 2017,
  Revised Selected Papers 3}, pages 474--485. Springer, 2018.

\bibitem[Klarner et~al.(2023)Klarner, Rudner, Reutlinger, Schindler, Morris,
  Deane, and Teh]{klarner2023qsavi}
Leo Klarner, Tim G.~J. Rudner, Michael Reutlinger, Torsten Schindler,
  Garrett~M. Morris, Charlotte Deane, and Yee~Whye Teh.
\newblock Drug discovery under covariate shift with domain-informed prior
  distributions over functions.
\newblock In \emph{Proceedings of the 40th International Conference on Machine
  Learning}, 2023.

\bibitem[Kompa et~al.(2021)Kompa, Snoek, and Beam]{kompa2021second}
Benjamin Kompa, Jasper Snoek, and Andrew~L Beam.
\newblock Second opinion needed: communicating uncertainty in medical machine
  learning.
\newblock \emph{NPJ Digital Medicine}, 4\penalty0 (1):\penalty0 4, 2021.

\bibitem[Liang et~al.(2023)Liang, Zadeh, and Morency]{foundationsmultimodal}
Paul~Pu Liang, Amir Zadeh, and Louis-Philippe Morency.
\newblock Foundations and trends in multimodal machine learning: Principles,
  challenges, and open questions, 2023.

\bibitem[MacKay(1992)]{mckay1992practical}
David J.~C. MacKay.
\newblock A practical {B}ayesian framework for backpropagation networks.
\newblock \emph{Neural Comput.}, 4\penalty0 (3):\penalty0 448–472, May 1992.
\newblock ISSN 0899-7667.
\newblock \doi{10.1162/neco.1992.4.3.448}.

\bibitem[Miller et~al.(2014)Miller, Ng, Eslick, Tong, and
  Chen]{miller2014advanced}
David~C Miller, Brenda Ng, John Eslick, Charles Tong, and Yang Chen.
\newblock Advanced computational tools for optimization and uncertainty
  quantification of carbon capture processes.
\newblock In \emph{Computer Aided Chemical Engineering}, volume~34, pages
  202--211. Elsevier, 2014.

\bibitem[Muhammad et~al.(2021)Muhammad, Alshehri, Karray, El~Saddik,
  Alsulaiman, and Falk]{muhammad2021comprehensive}
Ghulam Muhammad, Fatima Alshehri, Fakhri Karray, Abdulmotaleb El~Saddik,
  Mansour Alsulaiman, and Tiago~H Falk.
\newblock A comprehensive survey on multimodal medical signals fusion for smart
  healthcare systems.
\newblock \emph{Information Fusion}, 76:\penalty0 355--375, 2021.

\bibitem[Nado et~al.(2022)Nado, Band, Collier, Djolonga, Dusenberry, Farquhar,
  Feng, Filos, Havasi, Jenatton, Jerfel, Liu, Mariet, Nixon, Padhy, Ren,
  Rudner, Sbahi, Wen, Wenzel, Murphy, Sculley, Lakshminarayanan, Snoek, Gal,
  and Tran]{nado2022uncertainty}
Zachary Nado, Neil Band, Mark Collier, Josip Djolonga, Michael~W. Dusenberry,
  Sebastian Farquhar, Qixuan Feng, Angelos Filos, Marton Havasi, Rodolphe
  Jenatton, Ghassen Jerfel, Jeremiah Liu, Zelda Mariet, Jeremy Nixon, Shreyas
  Padhy, Jie Ren, Tim G.~J. Rudner, Faris Sbahi, Yeming Wen, Florian Wenzel,
  Kevin Murphy, D.~Sculley, Balaji Lakshminarayanan, Jasper Snoek, Yarin Gal,
  and Dustin Tran.
\newblock Uncertainty baselines: Benchmarks for uncertainty \& robustness in
  deep learning, 2022.

\bibitem[Neal(1996)]{neal1996bayesian}
Radford~M Neal.
\newblock {B}ayesian {L}earning for {N}eural {N}etworks.
\newblock 1996.

\bibitem[Ovadia et~al.(2019{\natexlab{a}})Ovadia, Fertig, Ren, Nado, Sculley,
  Nowozin, Dillon, Lakshminarayanan, and Snoek]{ovadia2019uncertainty}
Yaniv Ovadia, Emily Fertig, Jie Ren, Zachary Nado, D.~Sculley, Sebastian
  Nowozin, Joshua Dillon, Balaji Lakshminarayanan, and Jasper Snoek.
\newblock {C}an you trust your model's uncertainty? {E}valuating predictive
  uncertainty under dataset shift.
\newblock In \emph{Advances in Neural Information Processing Systems 32}.
  2019{\natexlab{a}}.

\bibitem[Ovadia et~al.(2019{\natexlab{b}})Ovadia, Fertig, Ren, Nado, Sculley,
  Nowozin, Dillon, Lakshminarayanan, and Snoek]{ovadia2019can}
Yaniv Ovadia, Emily Fertig, Jie Ren, Zachary Nado, David Sculley, Sebastian
  Nowozin, Joshua Dillon, Balaji Lakshminarayanan, and Jasper Snoek.
\newblock Can you trust your model's uncertainty? evaluating predictive
  uncertainty under dataset shift.
\newblock \emph{Advances in neural information processing systems}, 32,
  2019{\natexlab{b}}.

\bibitem[Paszke et~al.(2019)Paszke, Gross, Massa, Lerer, Bradbury, Chanan,
  Killeen, Lin, Gimelshein, Antiga, et~al.]{pytorch}
Adam Paszke, Sam Gross, Francisco Massa, Adam Lerer, James Bradbury, Gregory
  Chanan, Trevor Killeen, Zeming Lin, Natalia Gimelshein, Luca Antiga, et~al.
\newblock Pytorch: An imperative style, high-performance deep learning library.
\newblock \emph{Advances in neural information processing systems}, 32, 2019.

\bibitem[Rudner et~al.(2022{\natexlab{a}})Rudner, Chen, Teh, and Gal]{fsvi}
Tim G.~J. Rudner, Zonghao Chen, Yee~Whye Teh, and Yarin Gal.
\newblock Tractable function-space variational inference in {B}ayesian neural
  networks.
\newblock In \emph{Advances in Neural Information Processing Systems 35},
  2022{\natexlab{a}}.

\bibitem[Rudner et~al.(2022{\natexlab{b}})Rudner, Smith, Feng, Teh, and
  Gal]{rudner2022sfsvi}
Tim G.~J. Rudner, Freddie~Bickford Smith, Qixuan Feng, Yee~Whye Teh, and Yarin
  Gal.
\newblock Continual learning via sequential function-space variational
  inference.
\newblock In \emph{Proceedings of the 39th International Conference on Machine
  Learning}, 2022{\natexlab{b}}.

\bibitem[Rudner et~al.(2023{\natexlab{a}})Rudner, Kapoor, Qiu, and
  Wilson]{rudner2023fseb}
Tim G.~J. Rudner, Sanyam Kapoor, Shikai Qiu, and Andrew~Gordon Wilson.
\newblock Function-space regularization in neural networks: A probabilistic
  perspective.
\newblock In \emph{Proceedings of the 40th International Conference on Machine
  Learning}, Proceedings of Machine Learning Research. PMLR,
  2023{\natexlab{a}}.

\bibitem[Rudner et~al.(2023{\natexlab{b}})Rudner, Kapoor, Qiu, and
  Wilson]{rudner2023fsmap}
Tim G.~J. Rudner, Sanyam Kapoor, Shikai Qiu, and Andrew~Gordon Wilson.
\newblock Should we learn most likely functions or parameters?
\newblock In \emph{Advances in Neural Information Processing Systems 36},
  2023{\natexlab{b}}.

\bibitem[Shamout et~al.(2021)Shamout, Shen, Wu, Kaku, Park, Makino,
  Jastrzkeski, Witowski, Wang, Zhang, et~al.]{shamout2021artificial}
Farah~E Shamout, Yiqiu Shen, Nan Wu, Aakash Kaku, Jungkyu Park, Taro Makino,
  Stanislaw Jastrzkeski, Jan Witowski, Duo Wang, Ben Zhang, et~al.
\newblock An artificial intelligence system for predicting the deterioration of
  covid-19 patients in the emergency department.
\newblock \emph{NPJ digital medicine}, 4\penalty0 (1):\penalty0 80, 2021.

\bibitem[Tran et~al.(2022)Tran, Liu, Dusenberry, Phan, Collier, Ren, Han, Wang,
  Mariet, Hu, Band, Rudner, Singhal, Nado, van Amersfoort, Kirsch, Jenatton,
  Thain, Yuan, Buchanan, Murphy, Sculley, Gal, Ghahramani, Snoek, and
  Lakshminarayanan]{Plex22}
Dustin Tran, Jeremiah Liu, Michael~W. Dusenberry, Du~Phan, Mark Collier, Jie
  Ren, Kehang Han, Zi~Wang, Zelda Mariet, Huiyi Hu, Neil Band, Tim G.~J.
  Rudner, Karan Singhal, Zachary Nado, Joost van Amersfoort, Andreas Kirsch,
  Rodolphe Jenatton, Nithum Thain, Honglin Yuan, Kelly Buchanan, Kevin Murphy,
  D.~Sculley, Yarin Gal, Zoubin Ghahramani, Jasper Snoek, and Balaji
  Lakshminarayanan.
\newblock Plex: Towards reliability using pretrained large model extensions,
  2022.

\bibitem[Wainwright and Jordan(2008)]{wainwright2008vi}
Martin~J Wainwright and Michael~I Jordan.
\newblock \emph{Graphical Models, Exponential Families, and Variational
  Inference}.
\newblock Now Publishers Inc., Hanover, MA, USA, 2008.
\newblock ISBN 1601981848.

\bibitem[Xia et~al.(2022)Xia, Han, and Mascolo]{xia2022benchmarking}
Tong Xia, Jing Han, and Cecilia Mascolo.
\newblock Benchmarking uncertainty quantification on biosignal classification
  tasks under dataset shift.
\newblock In \emph{Multimodal AI in healthcare: A paradigm shift in health
  intelligence}, pages 347--359. Springer, 2022.

\bibitem[Zhang et~al.(2020)Zhang, Dong, Wang, Yu, Yao, Zhou, Hu, Li,
  Jim{\'e}nez-Mesa, Ramirez, et~al.]{zhang2020advances}
Yu-Dong Zhang, Zhengchao Dong, Shui-Hua Wang, Xiang Yu, Xujing Yao, Qinghua
  Zhou, Hua Hu, Min Li, Carmen Jim{\'e}nez-Mesa, Javier Ramirez, et~al.
\newblock Advances in multimodal data fusion in neuroimaging: overview,
  challenges, and novel orientation.
\newblock \emph{Information Fusion}, 64:\penalty0 149--187, 2020.

\bibitem[Zou et~al.(2023)Zou, Chen, Yuan, Shen, Wang, and Fu]{zou2023review}
Ke~Zou, Zhihao Chen, Xuedong Yuan, Xiaojing Shen, Meng Wang, and Huazhu Fu.
\newblock A review of uncertainty estimation and its application in medical
  imaging.
\newblock \emph{arXiv preprint arXiv:2302.08119}, 2023.

\end{thebibliography}

\clearpage

\begin{appendices}

\crefalias{section}{appsec}
\crefalias{subsection}{appsec}
\crefalias{subsubsection}{appsec}

\setcounter{equation}{0}
\renewcommand{\theequation}{\thesection.\arabic{equation}}

\onecolumn

\appendix

\vbox{%
\hsize\textwidth
\linewidth\hsize
\hrule height 4pt%
  \vskip 0.25in
  \vskip -\parskip%
\centering
{\LARGE\bf
Supplementary Material
\par}
\vskip 0.29in
  \vskip -\parskip
  \hrule height 1pt
  \vskip 0.09in%
}

 \label{appendix-a}
 \setcounter{table}{0}
\renewcommand{\thetable}{A\arabic{table}}

 \setcounter{figure}{0}
\renewcommand{\thefigure}{A\arabic{figure}}

\section{Variational Objective}
\label{appsec:variational_objective}

Let the mapping $f$ in the parametric observation model $p_{Y | X, \Theta}(y \vbar x, \theta; f)$ be defined by \mbox{$f(\cdot \,; \theta) \defines h(\cdot \,; \theta_{h}) \theta_{L}$}.
For a neural network model, $h(\cdot \,; \theta_{h})$ is the post-activation output of the penultimate layer, $\Theta_{L}$ is the set of stochastic final-layer parameters, $\Theta_{h}$ is the set of stochastic non-final-layer parameters, and $\Theta \defines \{ \Theta_{h} , \Theta_{L}\}$ is the full set of stochastic parameters.

To derive an uncertainty-aware prior distribution over the set of random parameters $\Theta$, we start by specifying an auxiliary inference problem.
Let \mbox{$\tilde{x} = \{ x_{1}, ..., x_{M} \}$} be a set of context points with corresponding labels $\tilde{y}$, and define a corresponding likelihood function $\tilde{p}_{Y | X, \Theta}(\tilde{y} \vbar \tilde{x} , \theta)$ and a prior over the model parameters, $p_{\Theta}(\theta)$.
For notational simplicity, we will drop the subscripts as we advance except when needed for clarity.
By Bayes' Theorem, we can write the posterior under the context points and labels as
\begin{align}
    \tilde{p}(\theta \vbar \tilde{x}, \tilde{y})
    \propto
    \tilde{p}(\tilde{y} \vbar \tilde{x} , \theta_{h}) p(\theta_{h}) p(\theta_{L}) .
\end{align}
To define a likelihood function that induces a posterior with desirable properties, we start from the same step as \citet{rudner2023fseb} and consider the following stochastic linear model for an arbitrary set of points $x \defines \{ x_{1}, ..., x_{M'} \}$,\vspace*{-3pt}
\begin{align*}
    \tilde{Y}_{k}(x)
    \defines
    h(x ; \theta_{h}) \Theta_{k} + \varepsilon
    \quad
    \text{with} ~~ \Theta_{k} \sim \calN(\theta_{L} ; m_{k}, \tau_{f}^{-1} s_{k}) ~~ \text{and} ~~  \varepsilon \sim \calN(\mathbf{0}, \tau_{f}^{-1}\beta I)
\end{align*}%
for output dimensions $k = 1, ..., K$, where $h(\cdot \,; \theta_{h})$ is the feature mapping used to define $f$ evaluated at a set of fixed feature parameters $\theta_{h}$, $\tau_{f}$ and $\beta$ are variance parameters, and $m \in \mathbb{R}^{P_{L}}$ and $s \in \mathbb{R}^{P_{L}}$ are---for now---fixed parameters for a $P_{L}$-dimensional final layer.
This stochastic linear model induces a distribution over functions~\citep{fsvi,rudner2022sfsvi,klarner2023qsavi,rudner2023fsmap}, which---when evaluated at $\tilde{x}$---is given by
\begin{align}
    \calN(\tilde{y}_{k}(\tilde{x}) ; h(\tilde{x} ; \theta_{h}) m_{k}, \tau_{f}^{-1} K(\tilde{x}, \tilde{x} ; \theta_{h}, s)_{k} ) ,
    \label{eq:induced_prior_distribution}
\end{align}\\[-17pt]
where\vspace*{-5pt}
\begin{align}
    K(\tilde{x}, \tilde{x} ; \theta_{h}, s)_{k}
    \defines
     h(\tilde{x} ; \theta_{h}) ( s_{k} I ) h(\tilde{x} ; \theta_{h})^\top + \beta I
    \label{eq:covariance}
\end{align}
is an $M$-by-$M$ covariance matrix.
Viewing this probability density over function evaluations as a likelihood function parameterized by $\theta$, we diverge from \citet{rudner2023fseb} and define
\begin{align}
\begin{split}
    \hspace*{-5pt}\tilde{p}(\tilde{y}_{k} \vbar \tilde{x} , \theta_{h})
    \hspace*{-1pt}\defines\hspace*{-1pt}
    \calN(\tilde{y}_{k} ; h(\tilde{x} ; \theta_{h}) m_{k} , \tau_{f}^{-1} K(\tilde{x}, \tilde{x} ; \theta_{h}, s)_{k} ) ,\hspace*{-3pt}
    \label{eq:aux_likelihood}
\end{split}
\end{align}
where---unlike in \citet{rudner2023fseb}---we do not assume that $m = \mathbf{0}$ and $s = I$.
If we define the auxiliary label distribution as $p_{\smash{\tilde{Y} \vbar \tilde{X}}}(\tilde{y} \vbar \tilde{x}) \defines \delta(\{\mathbf{0}, ..., \mathbf{0} \} - \tilde{y})$, the likelihood $\tilde{p}(\tilde{y}_{k} \vbar \tilde{x} , \theta_{h})$ favors learnable parameters $\theta_{h}$ for which the induced distribution over functions has a high likelihood of predicting $\mathbf{0}$.
Letting\vspace*{-5pt}
\begin{align*}
    \tilde{p}(\tilde{y} \vbar \tilde{x} , \theta)
    \defines
    \prod\nolimits_{k = 1}^{K} \tilde{p}(\tilde{y}_{k} \vbar \tilde{x} , \theta, m_{k}, s_{k}) ,
\end{align*}\\[-9pt]
and taking the log of the analytically tractable density $\tilde{p}(\tilde{y} \vbar \tilde{x} , \theta ; f)$, we obtain
\begin{align*}
    &
    \log \tilde{p}(\tilde{y} \vbar \tilde{x} , \theta_{h})
    \propto
    \hspace*{-2pt}-\hspace*{-2pt}
    \sum\nolimits_{k = 1}^{K} \frac{\tau_{f}}{2} (h(\tilde{x} ; \theta_{h}) m_{k})^\top K(\tilde{x}, \tilde{x} ; \theta_{h}, s)_{k}^{-1} h(\tilde{x} ; \theta_{h}) m_{k} ,
    \nonumber
\end{align*}
with proportionality up to an additive constant independent of $\theta$.
We define\vspace*{-2pt}
\begin{align}
\begin{split}
    &
    \hspace*{-5pt}\mathcal{J}(\theta, m, s, \tilde{x}, \tilde{y})
    \hspace*{-5pt}~
    \defines
    \hspace*{-2pt}-\hspace*{-2pt}
    \sum\nolimits_{k = 1}^{K}
    \frac{\tau_{f}}{2} d^{2}_{M}(h(\tilde{x} ; \theta_{h}) m_{k} - \tilde{y}, K(\tilde{x}, \tilde{x} ; \theta_{h}, s)_{k} ) 
    \label{eq:fs_map_regularizer}
\end{split}
\end{align}\\[-5pt]
where \mbox{$d^{2}_{M}(\Delta, K) \defines \Delta^\top K^{-1} \Delta$} is the squared Mahalanobis distance for $\Delta = v - w$.
We therefore obtain\vspace*{-2pt}
\begin{align*}
    \argmax\nolimits_{\theta} \tilde{p}(\theta \vbar \tilde{x}, \tilde{y})
    \hspace*{-2pt}=\hspace*{-2pt}
    \argmax\nolimits_{\theta}\mathcal{J}(\theta, m, s, \tilde{x}, \tilde{y})
    \hspace*{-2pt}+\hspace*{-2pt}
    \log p(\theta) 
\end{align*}\\[-13pt]
and hence, maximizing $\mathcal{J}(\theta, m, s, \tilde{x}, \tilde{y}) + \log p(\theta)$ with respect to $\theta$ is mathematically equivalent to maximizing the posterior $\tilde{p}(\theta \vbar \tilde{x}, \tilde{y})$ and leads to functions that are likely under the distribution over functions induced by the neural network mapping while being consistent with the prior over the network parameters.

However, since the parameters $m$ and $s$ are fixed and appear in the auxiliary likelihood function but not in the predictive function $f(\cdot \,; \theta)$, the objective above is not a good choice if the goal is to find parameters $\theta$ that induce functions that have high predictive uncertainty on the set of context points.
To address this shortcoming, we include these parameters in the observation model as the mean and variance parameters of the final-layer parameters in $f(\cdot \,; \theta)$, treat them as random variables $M$ and $S$, place a prior over them, and ultimately infer an approximate posterior distribution for both.

In particular, we define a prior over the final-layer parameters $\Theta_{L}$ as
\begin{align}
    p_{\Theta_{L}}(\theta_{L} \vbar m, s)
    =
    \calN(\theta_{L} ; m, s I)
\end{align}
and corresponding hyperpriors
\begin{align}
    p_{M}(m)
    &
    =
    \calN(m ; \mu_{0}, \tau_{0}^{-1} I)
    \\
    p_{S}(s)
    &
    =
    \textrm{Lognormal}(s ; \mathbf{0}, 2 \tau^{-1}_{s} I) .
\end{align}
As before, we will drop subscripts for brevity unless needed for clarity.
The full probabilistic model then becomes\vspace*{-5pt}
\begin{align}
    p(y \vbar x, \theta_{h}, \theta_{L}; f) \, p(\theta, m, s \vbar \tilde{x}, \tilde{y}) .
\end{align}
with the prior factorizing and simplifying as
\begin{align}
\begin{split}
    p(\theta, m, s \vbar \tilde{x}, \tilde{y})
    &
    =
    \tilde{p}(\theta_{h} \vbar m, s, \smash{\tilde{x}, \tilde{y}}) \, p(\theta_{L} \vbar m, s) \, p(m) \, p(s)
    \\
    &
    \propto
    p(\theta_{L} \vbar m, s) \, \tilde{p}(\tilde{y} \vbar \tilde{x} , \theta ; f) \, p(\theta_{h}) \, p(m) \, p(s) ,
\end{split}
\end{align}
all of which we can compute analytically.
With this prior, we can now derive a variational objective and perform approximate inference.

We begin by defining a variational distribution,
\begin{align*}
    q(\theta, m, s, \tilde{x}, \tilde{y})
    \defines
    q(\theta_{h}) \, q(\theta_{L} \vbar m, s) \, q(m) \, q(s) \, q(\tilde{x}, \tilde{y}) ,
\end{align*}
and frame the inference problem of finding the posterior $p(\theta, m, s, \tilde{x}, \tilde{y} \vbar x_{\calD}, y_{\calD})$ as a problem of optimization,
\begin{align*}
    \min_{q_{\Theta, M, S, \tilde{X}, \tilde{Y}} \in \calQ} \DKL{q_{\Theta, M, S, \tilde{X}, \tilde{Y}}}{p_{\Theta, M, S, \tilde{X}, \tilde{Y} \vbar X_{\calD}, Y_{\calD}}} ,
\end{align*}
where $\smash{\calQ}$ is a variational family.
If the posterior $\smash{p_{\Theta, M, S, \tilde{X}, \tilde{Y} | X_{\calD}, Y_{\calD}}}$ is in the variational family $\calQ$, then the solution to the variational minimization problem is equal to the exact posterior.
Modifying the inference problem by defining $q(\tilde{x}, \tilde{y}) \defines p(\tilde{x}, \tilde{y}) = p(\tilde{y} \vbar \tilde{x}) p(\tilde{x})$, which further constrains the variational family, the optimization problem simplifies to
\begin{align*}
    \min_{q_{\Theta, M, S} \in \calQ} \mathbb{E}_{p_{\tilde{X}, \tilde{Y}}} \left[ \DKL{q_{\Theta, M, S}}{p_{\Theta, M, S \vbar \smash{\tilde{X}, \tilde{Y}}, X_{\calD}, Y_{\calD}}} \right] ,
\end{align*}
which can equivalently be expressed as maximizing the variational objective
\begin{align*}
    \bar{\calF}(q_{\Theta}, q_{M}, q_{S})
    &
    \defines
    \mathbb{E}_{q_{\Theta, M, S}} [ \log p(y_{\calD} \vbar x_{\calD} , \Theta ; f) ]
    - \mathbb{E}_{p_{\tilde{X}, \tilde{Y}}} [ \DKL{q_{\Theta, M, S}}{p_{\Theta, M, S \vbar \smash{\tilde{X}, \tilde{Y}}}} ] .
\end{align*}
To obtain a tractable expression of the regularization term, we first note that we can write
\begin{align}
\begin{split}    
    \mathbb{E}_{p_{\tilde{X}, \tilde{Y}}} [ \DKL{q_{\Theta, M, S}}{p_{\Theta, M, S \vbar \smash{\tilde{X}, \tilde{Y}}}}] ]
    =
    \mathbb{E}_{p_{\tilde{X}, \tilde{Y}}} \Big[ \mathbb{E}_{q_{\Theta} q_{M} q_{S}} [ \log q(\Theta) q(M) q(S) ]
    - \mathbb{E}_{q_{\Theta} q_{M} q_{S}} [ \log p(\Theta, M, S \vbar \smash{\tilde{X}, \tilde{Y}}) ] \Big] ,
    \label{eq:kl_divergence}
\end{split}
\end{align}
where the first term is the negative entropy and the second term is the negative cross-entropy.
Using the same insights as above, we can write
\begin{align}
\begin{split}
    \mathbb{E}_{p_{\tilde{X}, \tilde{Y}}} [ \mathbb{E}_{q_{\Theta} q_{M} q_{S}} [ \log p(\Theta, M, S \vbar \smash{\tilde{X}, \tilde{Y}}) ] ]
    &
    \propto
    \mathbb{E}_{p_{\tilde{X}, \tilde{Y}}} \Big[\mathbb{E}_{q_{\Theta_{h}} q_{M} q_{S}} \left[ \log \tilde{p}(\smash{\tilde{Y} \vbar \tilde{X}} , \Theta_{h}, M, S) \right]
    \\
    &
    \qquad~~
    + \mathbb{E}_{q_{\Theta}} \left[ \log p(\Theta_{h}) \, p(\Theta_{L} \vbar M, S) \, p(M) \, p(S) \right] \Big] ,
\end{split}
\end{align}
up to an additive constant independent of $\theta$, and use it to express the \kld in \Cref{eq:kl_divergence} up to an additive constant independent of $\theta$ as
\begin{align}
\begin{split}
    \nonumber
    \DKL{q_{\Theta, M, S}}{p_{\Theta, M, S \vbar \smash{\tilde{X}, \tilde{Y}}}}
    &
    \propto
    -
    \mathbb{E}_{q_{M} q_{S}} [ \mathbb{E}_{q_{\Theta}} [\log \tilde{p}(\tilde{Y} \vbar \tilde{X} , \Theta_{h}, M, S) ]
    \\
    &
    \qquad
    +
    \DKL{q_{\Theta_{L} \vbar M, S}}{p_{\Theta_{L} \vbar M, S}} ]
    + \DKL{q_{\Theta_{h}}}{p_{\Theta_{h}}}
    \\
    &
    \qquad\qquad
    + \DKL{q_{M}}{p_{M}} + \DKL{q_{S}}{p_{S}} .
\end{split}
\end{align}
Now, further specifying the variational family as
\begin{align}
\begin{split}
    q(\theta_{L} \vbar m, s)
    &
    =
    \calN(\theta_{L} ; m, s I)
    \\
    q(\theta_{h})
    &
    =
    \calN(\theta_{h} ; \mu_{h}, \Sigma_{h})
    \\
    q(m)
    &
    =
    \calN(m ; \mu_{L}, \Sigma_{L})
    \\
    q(s)
    &
    =
    \textrm{Lognormal}(s ; \Sigma_{L}, \sigma^{2}_{s} I) .
\end{split}
\end{align}
with learnable variational parameters $\mu \defines \{ \mu_{h}, \mu_{m} \}$ and $ \Sigma \defines \{ \Sigma_{L}, \Sigma_{L} \}$ and fixed parameters $\{ \sigma^{2}_{m}, \sigma^{2}_{s} \}$, we get $\DKL{q_{\Theta_{L} \vbar M, S}}{p_{\Theta_{L} \vbar M, S}} = 0$, and the \kld simplifies to
\begin{align}
\begin{split}
    \DKL{q_{\Theta, M, S}}{p_{\Theta, M, S \vbar \smash{\tilde{X}, \tilde{Y}}}}
    &
    ~
    \propto
    -
    \mathbb{E}_{q_{\Theta_{h}} q_{M} q_{S}} [\log \tilde{p}(\tilde{Y} \vbar \tilde{X} , \Theta_{h}, M, S) ] + \DKL{q_{\Theta_{h}}}{p_{\Theta_{h}}}
    \\
    &
    \qquad
    + \DKL{q_{M}}{p_{M}} + \DKL{q_{S}}{p_{S}} ,
\end{split}
\end{align}
where each of the KL divergences can be computed analytically, and we can obtain an unbiased estimator of the negative log-likelihood using simple Monte Carlo estimation.

Since $\Theta_{h}$ and $q_{M}$ are both mean-field Gaussian distributions, we can equivalently express the full variational objective in a simplified form as
\begin{align}
\begin{split}
    \calF(\mu, \Sigma)
    &
    \defines
    \underbrace{\mathbb{E}_{q_{\Theta} q_{M} q_{S}} [ \log p(y_{\calD} \vbar x_{\calD} , \Theta ; f) ]}_{\textrm{Expected log-likelihood}} - \underbrace{\DKL{q_{\Phi}}{p_{\Phi}}}_{\textrm{KL regularization}}
    + \underbrace{ \mathbb{E}_{q_{\Theta_{h}} q_{M} q_{S}} [ \mathbb{E}_{p_{\tilde{X}, \tilde{Y}}} [ \log \tilde{p}(\smash{\tilde{Y} \vbar \tilde{X}} , \Theta_{h}, M, S) ]] - \smash{\tau_{s} \| \Sigma_{L} \|_{2}^{2}}}_{\textrm{Uncertainty regularization}}
    ,
\end{split}
\end{align}
where we defined $\Phi \defines \{ \Theta_{h}, M \}$.%
We can estimate the expectations in the objective using simple Monte Carlo estimation, and gradients can be estimated using reparameterization gradients as in \citet{blundell2015mfvi}.

Letting $p_{\tilde{Y} | \tilde{X}}(\tilde{y} \vbar \tilde{x}) = \delta(\mathbf{0}) $ to encourage high uncertainty in the predictions on the set of context points, where $\delta(\cdot)$ is the Dirac delta function, we obtain the simplified objective
\begin{align}
\begin{split}
    \calF(\mu, \Sigma)
    &
    \defines
    \underbrace{\mathbb{E}_{q_{\Theta} q_{M} q_{S}} [ \log p(y_{\calD} \vbar x_{\calD} , \Theta ; f) ]}_{\textrm{Expected log-likelihood}} - \underbrace{\DKL{q_{\Phi}}{p_{\Phi}}}_{\textrm{KL regularization}}
    + \underbrace{ \mathbb{E}_{q_{\Theta_{h}} q_{M} q_{S}} [ \mathbb{E}_{p_{\tilde{X}}} [ \log \tilde{p}(\smash{\mathbf{0} \vbar \tilde{X}} , \Theta_{h}, M, S) ]] - \smash{\tau_{s} \| \Sigma_{L} \|_{2}^{2}}}_{\textrm{Uncertainty regularization}}
    \label{eq:final_objective}
\end{split}
\end{align}

\clearpage

\section{Experimental details}
\label{app:experimental-setup}

\subsection{Training details}
For model training, we use the joint fusion protocol defined by \citet{medfuse} in which the network is trained end-to-end including the modality specific encoders $\Phi_{cxr}$ and $\Phi_{ehr}$ using the fully connected layer $g(\cdot)$ to obtain the multi-label probabilities ${\hat{y}}_{\textrm{fusion}}$.
\Cref{table:dataset_sizes} shows the details of the dataset splits used as input for our network.

\setlength{\tabcolsep}{21pt}
\begin{table}[h]
\caption{Summary of dataset sizes for the unimodal dataset and the combined multimodal dataset. We note that the size of the multimodal dataset decreases when the two modalities are paired.}
\small
\label{table:dataset_sizes}
\vspace*{-8pt}
    \centering

    \begin{tabular}{lcccc}
    \toprule
    \textbf{Dataset} & \textbf{Training} & \textbf{Validation} & \textbf{Testing} & \textbf{Context} \\
    \hline
    Clinical time series data  & 124,671 & 8,813 & 20,747 & 124,671 \\
    Chest X-rays & 42,628 & 4,802 & 11,914 & 42,628 \\
    Multimodal & 7,756 & 877 & 2,161 & 7,756 \\
    \bottomrule
    \end{tabular}
\end{table}

We use the binary cross-entropy loss \citep{good1952rational}, adapted to the multi-label classification task:
\begin{equation}
\label{equation:loss}
    \log p(y | x, \theta ; f)
    =
    -\sum_{i=1}^{n}(y_i\log(\hat{y}_i)+(1-y_i)(\log(1-\hat{y}_i))) ,
\end{equation}
where $\hat{y}_{i} \defines \textrm{sigmoid}(f(x_{i} ; \theta))$. The overall variational objective in our method is given by an expected log-likelihood term, KL regularization, and uncertainty regularization. In the stochastic setting, as described in \Cref{fig:fusion_encoder}, we combine the training and context datasets as the input for the computation of this loss.

\subsection{Hyperparameter tuning}
Initially, we used the deterministic baseline model to randomly sample a learning rate between $10^{-5}$ and $10^{-3}$ and selected the model and learning rate that achieved the model checkpoint with the best AUROC on the respective validation set. 
The best learning rate obtained was $2\times10^{-4}$, validated over 10 random seeds of training the deterministic model.

For the stochastic model, we performed a standard grid-based search to obtain the best hyper parameters for the regularization function. \Cref{table:hyperparams} shows the value ranges for each hyperparameter of our grid, which consists of 324 different model combinations.
\track{ 
We note that this procedure requires more resources due to the higher number of hyperparameters as compared to the deterministic model.
In addition, stochastic models also have more learnable parameters.
In our case two times as many parameters, since the model has mean and variance parameters, and the regularization term requires performing a forward pass on the number of context points sampled from the context distribution (which we choose to be fewer points than are contained in each minibatch).
In total, as is the case with every mean-field variational distribution, we have more learnable parameters than in a deterministic neural network and require more forward passes for every gradient step.}

\setlength{\tabcolsep}{34pt}
\begin{table}[h]
\caption{Hyperparameter grid search values for the stochastic model}
\small
\label{table:hyperparams}
    \centering
    \begin{tabular}{llc}
    \toprule
    \textbf{Hyperparameter} & \textbf{Values} & \textbf{Best} \\
    
    \hline
    
    prior variance & [1, 0.1, 0.01] & \textbf{0.1}\\
    prior likelihood scale & [1, 0.1, 10] & \textbf{1}\\
    prior likelihood f-scale & [0, 1, 10] & \textbf{10}\\
    prior likelihood covariance scale & [0.1, 0.01, 0.001, 0.0001] & \textbf{0.1}\\
    prior likelihood covariance diagonal & [1, 5, 0.5] & \textbf{5}\\
    
    \bottomrule
    \end{tabular}
\end{table}

\subsection{Model selection}
We trained our stochastic model for 400 epochs. 
Since we have four metrics of interest (i.e., AUROC, AUPRC, Selective AUROC and Selective AUPRC), we computed the hypervolume using the volume formula of a 4-dimensional sphere as the main aggregated metric to select the best model checkpoint during training.
\begin{equation}
\textrm{hypervolume}=\frac{\pi^2\textrm{R}^4}{2}
\end{equation}
where $\textrm{R}$ is the Euclidian magnitude of a 4-dimensional vector.
The hypervolume approach ensures that we do not overfit to a single metric in the process of finding the best model.

\subsection{Technical implementation}

Our data loading and pre-processing pipeline was implemented using PyTorch \citep{pytorch} following the same structure of the code used by \cite{medfuse}.
However, we refactored the original unimodal and multimodal models, training, and evaluation loops using JAX \citep{jax}.
This framework simplifies the implementation of Bayesian neural networks and stochastic training, which are the basis of the uncertainty quantification methods used in this work. 
In addition, we obtained a significant reduction in total training time for the unimodal and multimodal models using JAX, compared to PyTorch.

We note that due to specific caching procedures of the JAX framework, we had to standardize each $x_{\textrm{ehr}}$ instance into 300 time steps for the LSTM encoder to avoid out-of-memory issues.
The JAX framework requires that an LSTM encoder defines a static length of the sequences it is going to process, and then it caches this model in order to increase the training speed.
This means that if different sequence lengths are used, then JAX would cache an instance of the LSTM encoder for each specific length to be used during each training cycle.
The problem arises when dealing with a dataset that contains sequences of dynamic lengths that present high variance, i.e many different sequence lengths for every datapoint in the dataset, just as is the case with MIMIC-IV \citep{mimiciv}.
In comparison, PyTorch does not use this approach and is able to process sequences of dynamic length with one single instance of the LSTM encoder, however this is done at the cost of training speed when you compare both frameworks.

All of the experiments were executed using NVIDIA A100 and V100 80Gb Tensor Core GPUs. 

\newpage
\section{Additional Experimental Results}
\label{appsec:results}
In this section, we provide additional results on the test set. \Cref{table:batch_context_experiments} presents the results of the stochastic model for different context batch size values.

\setlength{\tabcolsep}{9pt}
\begin{table}[h]
\caption{Performance results on the test set for the stochastic model for varying values for context batch size.}
\small
\vspace*{-8pt}
\label{table:batch_context_experiments}
    \centering
    \begin{tabular}{ccccc}
    \toprule
    \textbf{\makecell{Context \\ batch size}} & \textbf{AUROC} & \textbf{AUPRC} & \textbf{\makecell{Selective \\ AUROC}} & \textbf{\makecell{Selective \\ AUPRC}}\\
    
    \hline

    16 & 0.732 (0.725, 0.739) & 0.511 (0.502, 0.525) & 0.740 (0.728, 0.753)	& 0.447 (0.432, 0.469) \\
    
    32 & 0.733 (0.725, 0.739) & 0.510 (0.500, 0.524) & 0.743 (0.733, 0.756)	& 0.448 (0.435, 0.466) \\
    
    64 & \cellcolor[gray]{0.9}\textbf{0.735} (0.728, 0.742) & \cellcolor[gray]{0.9}\textbf{0.514} (0.504, 0.528) & \cellcolor[gray]{0.9}\textbf{0.748} (0.738, 0.760) & \cellcolor[gray]{0.9}\textbf{0.452} (0.441, 0.472) \\
    
    128 & 0.733 (0.726, 0.739) & 0.512 (0.502, 0.525) & 0.728 (0.718, 0.739)	& 0.401 (0.391, 0.418) \\
    
    \bottomrule
    \end{tabular}
\end{table}

\Cref{table:per_label_deterministic}, \Cref{table:per_label_mfvi} and \Cref{table:per_label_bnn} present the extended results of our experiments for each label for the deterministic baseline, the Bayesian model with standard prior and the Bayesian model with \textsc{m2d2} prior, respectively.

\begin{table*}[ht]
\caption{Performance results across the different labels on the test set for the deterministic baseline \citep{medfuse}.}
\vspace*{-8pt}
\label{table:per_label_deterministic}
    \centering
    \resizebox{1.0\linewidth}{!}{
    \begin{tabular}{lccccc}
    \toprule
    \textbf{\makecell{Label}} &
    \textbf{Prevalence} & \textbf{AUROC} & \textbf{AUPRC} & \textbf{\makecell{Selective \\ AUROC}} & \textbf{\makecell{Selective \\ AUPRC}} \\
    \hline
    
    1 Acute and unspecified renal failure& 0.321 & 0.753 (0.732, 0.774) & 0.574 (0.537, 0.613) & 0.775 (0.750, 0.802) & 0.655 (0.600, 0.717) \\

    2 Acute cerebrovascular disease& 0.078 & 0.854 (0.819, 0.888) & 0.427 (0.353, 0.512) & 0.595 (0.526, 0.687) & 0.072 (0.066, 0.098) \\
    
    3 Acute myocardial infarction& 0.093 & 0.692 (0.654, 0.727) & 0.189 (0.153, 0.234) & 0.634 (0.585, 0.675) & 0.089 (0.080, 0.107) \\
    
    4 Cardiac dysrhythmias& 0.379 & 0.690 (0.668, 0.713) & 0.563 (0.530, 0.599) & 0.678 (0.649, 0.705) & 0.621 (0.573, 0.682) \\
    
    5 Chronic kidney disease& 0.240 & 0.755 (0.732, 0.779) & 0.495 (0.454, 0.541) & 0.855 (0.830, 0.878) & 0.600 (0.529, 0.689) \\
    
    6 Chronic obstructive pulmonary disease& 0.148 & 0.735 (0.706, 0.765) & 0.338 (0.293, 0.390) & 0.814 (0.776, 0.850) & 0.431 (0.334, 0.536) \\
    
    7 Complications of surgical/medical care& 0.226 & 0.677 (0.650, 0.704) & 0.382 (0.339, 0.428) & 0.583 (0.552, 0.621) & 0.212 (0.202, 0.243) \\
    
    8 Conduction disorders& 0.115 & 0.787 (0.750, 0.822) & 0.575 (0.515, 0.636) & 0.849 (0.814, 0.882) & 0.746 (0.691, 0.800) \\
    
    9 Congestive heart failure; nonhypertensive& 0.295 & 0.772 (0.750, 0.794) & 0.593 (0.554, 0.632) & 0.829 (0.808, 0.853) & 0.705 (0.652, 0.758) \\
    
    10 Coronary atherosclerosis and related& 0.337 & 0.764 (0.744, 0.784) & 0.624 (0.583, 0.663) & 0.842 (0.814, 0.866) & 0.700 (0.640, 0.762) \\
    
    11 Diabetes mellitus with complications& 0.120 & 0.848 (0.823, 0.872) & 0.485 (0.417, 0.552) & 0.757 (0.704, 0.831) & 0.250 (0.205, 0.305) \\
    
    12 Diabetes mellitus without complication& 0.211 & 0.710 (0.684, 0.737) & 0.359 (0.324, 0.402) & 0.680 (0.645, 0.731) & 0.251 (0.219, 0.306) \\
    
    13 Disorders of lipid metabolism& 0.406 & 0.694 (0.671, 0.715) & 0.593 (0.556, 0.630) & 0.749 (0.721, 0.776) & 0.671 (0.617, 0.720) \\
    
    14 Essential hypertension& 0.433 & 0.653 (0.630, 0.676) & 0.561 (0.525, 0.595) & 0.624 (0.598, 0.653) & 0.599 (0.549, 0.653) \\
    
    15 Fluid and electrolyte disorders& 0.454 & 0.711 (0.689, 0.731) & 0.681 (0.649, 0.713) & 0.688 (0.664, 0.713) & 0.779 (0.746, 0.811) \\
    
    16 Gastrointestinal hemorrhage& 0.071 & 0.629 (0.583, 0.677) & 0.135 (0.100, 0.183) & 0.590 (0.542, 0.646) & 0.078 (0.066, 0.091) \\
    
    17 Hypertension with complications& 0.222 & 0.746 (0.720, 0.768) & 0.452 (0.407, 0.499) & 0.842 (0.810, 0.868) & 0.549 (0.451, 0.644) \\
    
    18 Other liver diseases& 0.169 & 0.684 (0.654, 0.715) & 0.336 (0.291, 0.385) & 0.701 (0.642, 0.754) & 0.375 (0.286, 0.476) \\
    
    19 Other lower respiratory disease& 0.126 & 0.615 (0.580, 0.651) & 0.209 (0.172, 0.256) & 0.577 (0.539, 0.612) & 0.118 (0.109, 0.141) \\
    
    20 Other upper respiratory disease& 0.054 & 0.638 (0.585, 0.686) & 0.092 (0.068, 0.127) & 0.551 (0.489, 0.640) & 0.055 (0.049, 0.074) \\
    
    21 Pleurisy; pneumothorax; pulmonary& 0.095 & 0.665 (0.629, 0.698) & 0.182 (0.146, 0.230) & 0.607 (0.558, 0.658) & 0.088 (0.084, 0.111) \\
    
    22 Pneumonia& 0.185 & 0.733 (0.707, 0.758) & 0.373 (0.327, 0.427) & 0.739 (0.698, 0.787) & 0.316 (0.261, 0.407) \\
    
    23 Respiratory failure; insufficiency;& 0.282 & 0.786 (0.766, 0.807) & 0.603 (0.566, 0.642) & 0.841 (0.817, 0.864) & 0.719 (0.671, 0.769) \\
    
    24 Septicemia (except in labor)& 0.227 & 0.755 (0.731, 0.778) & 0.504 (0.460, 0.550) & 0.841 (0.809, 0.869) & 0.626 (0.558, 0.696) \\
    
    25 Shock& 0.174 & 0.816 (0.791, 0.840) & 0.554 (0.507, 0.604) & 0.867 (0.821, 0.903) & 0.663 (0.580, 0.728) \\

    \bottomrule
    \end{tabular}
    }
\end{table*}

\begin{table*}[ht]
\caption{Performance results across the different labels on the test set for the Bayesian (standard prior) model.}
\vspace*{-8pt}
\label{table:per_label_mfvi}
    \centering
    \resizebox{1.0\linewidth}{!}{
    \begin{tabular}{lccccc}
    \toprule
    \textbf{\makecell{Label}} &
    \textbf{Prevalence} & \textbf{AUROC} & \textbf{AUPRC} & \textbf{\makecell{Selective \\ AUROC}} & \textbf{\makecell{Selective \\ AUPRC}} \\
    \hline
    
    1 Acute and unspecified renal failure & 0.321 & 0.747 (0.726, 0.768) & 0.573 (0.537, 0.611) & 0.817 (0.790, 0.845) & 0.672 (0.618, 0.732) \\
    
    2 Acute cerebrovascular disease & 0.078 & 0.861 (0.828, 0.893) & 0.418 (0.350, 0.498) & 0.590 (0.520, 0.678) & 0.074 (0.067, 0.101) \\
    
    3 Acute myocardial infarction & 0.093 & 0.705 (0.670, 0.741) & 0.208 (0.166, 0.266) & 0.694 (0.656, 0.729) & 0.087 (0.078, 0.104) \\
    
    4 Cardiac dysrhythmias & 0.379 & 0.698 (0.676, 0.721) & 0.577 (0.544, 0.615) & 0.744 (0.707, 0.776) & 0.626 (0.570, 0.688) \\
    
    5 Chronic kidney disease & 0.240 & 0.739 (0.716, 0.762) & 0.476 (0.432, 0.520) & 0.808 (0.772, 0.841) & 0.574 (0.499, 0.648) \\
    
    6 Chronic obstructive pulmonary disease & 0.148 & 0.729 (0.700, 0.761) & 0.338 (0.290, 0.393) & 0.802 (0.762, 0.842) & 0.428 (0.334, 0.535) \\
    
    7 Complications of surgical/medical care & 0.226 & 0.657 (0.627, 0.685) & 0.383 (0.338, 0.429) & 0.672 (0.627, 0.716) & 0.394 (0.321, 0.473) \\
    
    8 Conduction disorders & 0.115 & 0.817 (0.781, 0.848) & 0.596 (0.537, 0.657) & 0.871 (0.832, 0.906) & 0.734 (0.650, 0.810) \\
    
    9 Congestive heart failure; nonhypertensive & 0.295 & 0.783 (0.762, 0.805) & 0.618 (0.579, 0.656) & 0.867 (0.840, 0.892) & 0.739 (0.688, 0.787) \\
    
    10 Coronary atherosclerosis and related & 0.337 & 0.766 (0.745, 0.787) & 0.645 (0.605, 0.680) & 0.836 (0.802, 0.862) & 0.735 (0.686, 0.781) \\
    
    11 Diabetes mellitus with complications & 0.120 & 0.811 (0.783, 0.837) & 0.429 (0.368, 0.495) & 0.844 (0.804, 0.874) & 0.359 (0.240, 0.444) \\
    
    12 Diabetes mellitus without complication & 0.211 & 0.691 (0.664, 0.717) & 0.354 (0.319, 0.399) & 0.642 (0.607, 0.682) & 0.222 (0.201, 0.265) \\
    
    13 Disorders of lipid metabolism & 0.406 & 0.685 (0.662, 0.708) & 0.579 (0.545, 0.614) & 0.759 (0.730, 0.785) & 0.617 (0.564, 0.674) \\
    
    14 Essential hypertension & 0.433 & 0.660 (0.637, 0.681) & 0.576 (0.543, 0.611) & 0.716 (0.684, 0.746) & 0.632 (0.582, 0.680) \\
    
    15 Fluid and electrolyte disorders & 0.454 & 0.714 (0.693, 0.734) & 0.668 (0.635, 0.700) & 0.708 (0.685, 0.736) & 0.756 (0.718, 0.796) \\
    
    16 Gastrointestinal hemorrhage & 0.071 & 0.638 (0.593, 0.682) & 0.131 (0.097, 0.177) & 0.606 (0.561, 0.676) & 0.071 (0.066, 0.092) \\
    
    17 Hypertension with complications & 0.222 & 0.733 (0.710, 0.757) & 0.431 (0.388, 0.479) & 0.779 (0.735, 0.818) & 0.486 (0.397, 0.579) \\
    
    18 Other liver diseases & 0.169 & 0.696 (0.664, 0.727) & 0.354 (0.305, 0.403) & 0.731 (0.680, 0.780) & 0.429 (0.345, 0.526) \\
    
    19 Other lower respiratory disease & 0.126 & 0.604 (0.567, 0.642) & 0.181 (0.153, 0.219) & 0.582 (0.547, 0.617) & 0.123 (0.113, 0.145) \\
    
    20 Other upper respiratory disease & 0.054 & 0.685 (0.632, 0.739) & 0.165 (0.109, 0.236) & 0.618 (0.552, 0.685) & 0.101 (0.052, 0.205) \\
    
    21 Pleurisy; pneumothorax; pulmonary & 0.095 & 0.666 (0.629, 0.701) & 0.166 (0.135, 0.208) & 0.593 (0.543, 0.657) & 0.090 (0.082, 0.120) \\
    
    22 Pneumonia & 0.185 & 0.758 (0.732, 0.781) & 0.400 (0.355, 0.455) & 0.783 (0.743, 0.829) & 0.341 (0.270, 0.436) \\
    
    23 Respiratory failure; insufficiency; & 0.282 & 0.824 (0.804, 0.843) & 0.631 (0.592, 0.675) & 0.890 (0.868, 0.909) & 0.703 (0.637, 0.771) \\
    
    24 Septicemia (except in labor) & 0.227 & 0.783 (0.761, 0.805) & 0.522 (0.476, 0.572) & 0.866 (0.834, 0.892) & 0.629 (0.530, 0.703) \\
    
    25 Shock & 0.174 & 0.826 (0.804, 0.847) & 0.552 (0.502, 0.606) & 0.888 (0.858, 0.918) & 0.582 (0.507, 0.658) \\
    
    \bottomrule
    \end{tabular}
    }
\end{table*}

\begin{table*}[ht]
\caption{Performance results across the different labels on the test set for the Bayesian (\textsc{m2d2} prior) model.}
\vspace*{-8pt}
\label{table:per_label_bnn}
    \centering
    \resizebox{1.0\linewidth}{!}{
    \begin{tabular}{lccccc}
    \toprule
    \textbf{\makecell{Label}} &
    \textbf{Prevalence} & \textbf{AUROC} & \textbf{AUPRC} & \textbf{\makecell{Selective \\ AUROC}} & \textbf{\makecell{Selective \\ AUPRC}} \\
    \hline

    1 Acute and unspecified renal failure & 0.321 & 0.756 (0.735, 0.779) & 0.587 (0.551, 0.627) & 0.830 (0.800, 0.857) & 0.680 (0.617, 0.740) \\

    2 Acute cerebrovascular disease & 0.078 & 0.870 (0.840, 0.901) & 0.459 (0.385, 0.545) & 0.664 (0.598, 0.745) & 0.088 (0.076, 0.115) \\
    
    3 Acute myocardial infarction & 0.093 & 0.716 (0.681, 0.754) & 0.220 (0.174, 0.277) & 0.656 (0.611, 0.698) & 0.083 (0.077, 0.104) \\
    
    4 Cardiac dysrhythmias & 0.379 & 0.687 (0.663, 0.710) & 0.570 (0.536, 0.606) & 0.737 (0.706, 0.768) & 0.654 (0.603, 0.709) \\
    
    5 Chronic kidney disease & 0.240 & 0.767 (0.744, 0.789) & 0.507 (0.464, 0.553) & 0.853 (0.825, 0.879) & 0.612 (0.534, 0.690) \\
    
    6 Chronic obstructive pulmonary disease & 0.148 & 0.727 (0.700, 0.757) & 0.326 (0.280, 0.377) & 0.773 (0.725, 0.814) & 0.413 (0.293, 0.503) \\
    
    7 Complications of surgical/medical care & 0.226 & 0.659 (0.631, 0.686) & 0.396 (0.351, 0.444) & 0.666 (0.620, 0.708) & 0.375 (0.310, 0.462) \\
    
    8 Conduction disorders & 0.115 & 0.798 (0.763, 0.832) & 0.593 (0.532, 0.656) & 0.850 (0.814, 0.889) & 0.754 (0.689, 0.812) \\
    
    9 Congestive heart failure; nonhypertensive & 0.295 & 0.788 (0.768, 0.808) & 0.600 (0.562, 0.637) & 0.874 (0.853, 0.892) & 0.722 (0.663, 0.773) \\
    
    10 Coronary atherosclerosis and related & 0.337 & 0.767 (0.746, 0.787) & 0.626 (0.588, 0.665) & 0.846 (0.820, 0.869) & 0.716 (0.654, 0.769) \\
    
    11 Diabetes mellitus with complications & 0.120 & 0.842 (0.817, 0.866) & 0.461 (0.398, 0.528) & 0.821 (0.765, 0.906) & 0.344 (0.256, 0.472) \\
    
    12 Diabetes mellitus without complication & 0.211 & 0.716 (0.690, 0.742) & 0.386 (0.345, 0.432) & 0.673 (0.634, 0.727) & 0.256 (0.223, 0.313) \\
    
    13 Disorders of lipid metabolism & 0.406 & 0.698 (0.675, 0.720) & 0.591 (0.558, 0.628) & 0.773 (0.746, 0.798) & 0.624 (0.572, 0.685) \\
    
    14 Essential hypertension & 0.433 & 0.669 (0.646, 0.694) & 0.588 (0.555, 0.621) & 0.709 (0.677, 0.740) & 0.643 (0.588, 0.695) \\
    
    15 Fluid and electrolyte disorders & 0.454 & 0.717 (0.695, 0.737) & 0.679 (0.647, 0.709) & 0.739 (0.712, 0.767) & 0.771 (0.732, 0.811) \\
    
    16 Gastrointestinal hemorrhage & 0.071 & 0.667 (0.627, 0.708) & 0.134 (0.104, 0.185) & 0.596 (0.537, 0.720) & 0.068 (0.064, 0.089) \\
    
    17 Hypertension with complications & 0.222 & 0.760 (0.736, 0.781) & 0.475 (0.430, 0.522) & 0.843 (0.810, 0.872) & 0.574 (0.483, 0.666) \\
    
    18 Other liver diseases & 0.169 & 0.723 (0.693, 0.750) & 0.378 (0.330, 0.428) & 0.673 (0.614, 0.743) & 0.313 (0.228, 0.423) \\
    
    19 Other lower respiratory disease & 0.126 & 0.594 (0.560, 0.629) & 0.181 (0.154, 0.222) & 0.568 (0.533, 0.612) & 0.120 (0.114, 0.144) \\
    
    20 Other upper respiratory disease & 0.054 & 0.670 (0.614, 0.722) & 0.133 (0.094, 0.193) & 0.540 (0.483, 0.603) & 0.052 (0.048, 0.071) \\
    
    21 Pleurisy; pneumothorax; pulmonary & 0.095 & 0.692 (0.658, 0.726) & 0.167 (0.139, 0.207) & 0.630 (0.576, 0.707) & 0.086 (0.081, 0.109) \\
    
    22 Pneumonia & 0.185 & 0.756 (0.732, 0.781) & 0.411 (0.362, 0.468) & 0.808 (0.769, 0.851) & 0.395 (0.328, 0.499) \\
    
    23 Respiratory failure; insufficiency; & 0.282 & 0.811 (0.790, 0.831) & 0.633 (0.594, 0.673) & 0.876 (0.848, 0.899) & 0.733 (0.675, 0.791) \\
    
    24 Septicemia (except in labor) & 0.227 & 0.774 (0.751, 0.797) & 0.513 (0.467, 0.559) & 0.818 (0.772, 0.865) & 0.582 (0.504, 0.664) \\
    
    25 Shock & 0.174 & 0.809 (0.786, 0.833) & 0.536 (0.486, 0.586) & 0.876 (0.838, 0.902) & 0.637 (0.546, 0.702) \\
    
    \bottomrule
    \end{tabular}
    }
\end{table*}

\end{appendices}

\end{document}